\documentclass[a4paper,fleqn]{cas-sc}

\usepackage[authoryear,longnamesfirst]{natbib}
\usepackage{subcaption}

\def\tsc#1{\csdef{#1}{\textsc{\lowercase{#1}}\xspace}}
\tsc{WGM}
\tsc{QE}
\tsc{EP}
\tsc{PMS}
\tsc{BEC}
\tsc{DE}

\begin{document}
\let\WriteBookmarks\relax
\def\floatpagepagefraction{1}
\def\textpagefraction{.001}
\shorttitle{Neural network approach to reservoir simulation and adaptation}
\shortauthors{Illarionov et~al.}

\title [mode = title]{End-to-end neural network approach to 3D reservoir simulation and adaptation}                      
\author[1]{Illarionov~E.}
\cormark[1]
\ead{e.illarionov@skoltech.ru}
\author[1]{Temirchev~P.}
\author[1]{Voloskov~D.}
\author[1]{Kostoev~R.}
\author[2]{Simonov~M.}
\author[1,3]{Pissarenko~D.}
\author[1]{Orlov~D.}
\author[1]{Koroteev~D.}
\address[1]{Skolkovo Institute of Science and Technology, Moscow, Russia}
\address[2]{LLC "Gazpromneft Science and Technology Centre", Saint-Petersburg, Russia}
\address[3]{TOTAL Research and Development, Moscow, Russia}

\cortext[cor1]{Corresponding author}

\begin{abstract}
Reservoir simulation and adaptation (also known as history matching) are typically considered as separate problems. While a set of models are
aimed at the solution of the forward simulation problem assuming
all initial geological parameters are known, the other set of
models adjust geological parameters under the fixed forward simulation model to fit production data. This results in many difficulties for both reservoir engineers and developers of new efficient computation schemes. We present a unified approach to reservoir simulation and adaptation problems. A single neural network model allows a forward pass from initial geological parameters of the 3D reservoir model through dynamic state variables to well's production rates and backward gradient propagation to any model inputs and variables.
The model fitting and geological parameters adaptation both become the optimization problem over specific parts of the same neural network model. Standard gradient-based optimization schemes can be used to find the optimal solution. Using real-world oilfield model and historical production rates we demonstrate that the suggested approach allows reservoir simulation and history matching with a benefit of several orders of magnitude simulation speed-up. Finally, to propagate this research we open-source a Python-based framework DeepField that allows standard processing of reservoir models and reproducing the approach presented in this paper.

\end{abstract}


\maketitle

\section{Introduction}

Reservoir simulation is a complex concept that typically includes 
a lot of steps ranging from construction of appropriate geological
model to an estimation of field performance, e.g. oil production rates.
However, once all the geological parameters (initial and boundary
conditions) are set, and control parameters are given, the challenge
is to simulate reservoir dynamics, i.e. estimate time-dependent
variables (phase saturation, pressure, well's production rates, etc.).
A standard approach is based on a system of hydrodynamic equations
and its numerical evaluation (see, e.g. \cite{chen_computational_2006}). While the physical equations can be assumed as fixed, numerical solution methods become
a matter of intense research.

Straightforward implementation of finite-difference
methods (i.e. discretization of physical equations)
guaranties to provide a solution, but requires 
enormous computational costs in practical cases.
Since the middle of the last century, investigations in
parallelized solution schemes or more advanced 
computation algorithms in application to the petroleum industry
become a separate research field. 
Since recently, implementation of machine learning methods into classical schemes is of special interest (see, e.g. \cite{ML_reservoir2020} for implementation details and \cite{KOROTEEV2021100041} for future perspectives).

Modern reservoir simulation schemes provide an excellent approximation
of field dynamics in synthetic cases \citep{Kvashchuk2019}. However, due to
natural uncertainties in the estimation of geological parameters
in real-world fields, uncertainties in obtained solutions
easily make them impractical. A process of adaptation in the space of initial and boundary conditions to align
simulated data and actual production rates is known as
history matching (HM). The problem is clearly ill-posed
and thus assumes various adaptation strategies
(see, e.g. \cite{Oliver2011} for a review).

Mathematically, HM is an optimization problem
over a certain space of variables. The point is that
simulated data are usually obtained with black-box
simulation tools that restrict straightforward
application of standard gradient-based optimization methods. 
Moreover, forward simulation usually requires high
computational and time costs, and it complicates
dramatically alternative approaches. As a result,
HM has become a separate research field.

One could substantially benefit
from considering reservoir simulation and adaptation
within a single framework. We present an approach
that makes it possible and, moreover, uses the same
optimization methods for the solution of both problems. We implement an end-to-end neural network
model for reservoir simulation and production
rates calculation. The model training
phase can be considered as an optimization
problem in the space of neural network variables given a
dataset of simulated field scenarios. In the
same way, given actual production rates,
we consider HM as an optimization
problem in the space of geological parameters.

By construction, neural network models allow
gradient backpropagation to any input and
internal variables
and one can apply common optimization algorithms
for model fitting (see \cite{Goodfellow2016} for general theory). Of course, 
gradient-based HM 
is not a new idea (see e.g. \cite{Kaleta} or \cite{Gomez}). The point is that in contrast to
previous works, estimation of gradients does not
require elaboration of separate models or substantial model
reduction. Gradients in neural network models can
be computed analytically and thanks to modern
programming frameworks, there is no need to do it
explicitly. Using a real-world oilfield model we
demonstrate that this approach provides accurate
solutions in reservoir simulation and adaptation
with several orders of magnitude speed benefit
in comparison to standard industrial software.
Note that in previous work \cite{Illarionov20203DRM} the HM problem was investigated only with respect to simulated
dynamic state variables (pressures and
phase saturations), while in this work, we consider
the most practical problem given historical
well's production rates.

Of course, the suggested
model is not aimed at the direct substitution of standard simulation software, which is based on finite-difference methods. While the last ones provide highly precise solutions at the cost of large computation time, the neural network approach allows faster approximation by means of reduced accuracy to some extent, which is expected in any proxy model.

\section{Dynamics module}

\begin{figure}[h]
\centering
\includegraphics[width=0.85\textwidth]{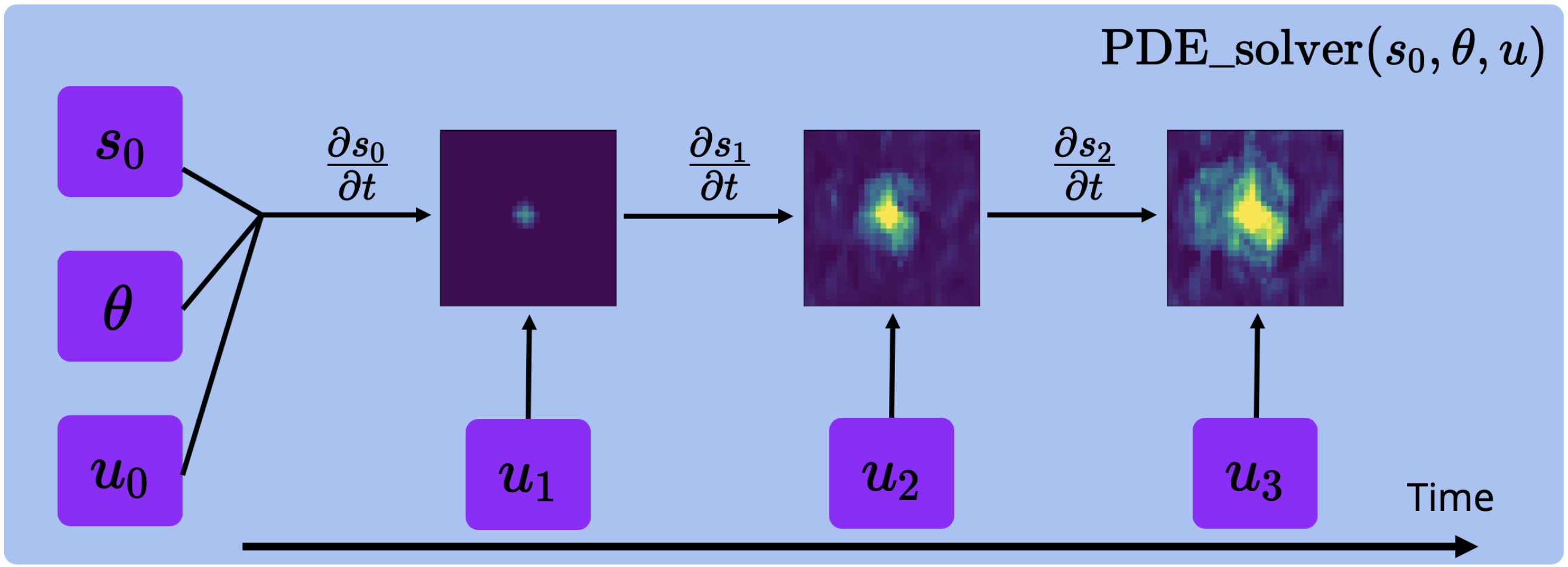}
\caption{Classical simulation scheme with the finite-differences method. Here $\theta$, $s_0$ and $u_{i}$ denote reservoir static variables, initial state and control parameters for the time interval. At each timestep the simulation process (\ref{simulation}) outputs the solution $s_i$ and we show only some slices of the 3D cubes obtained.}
\label{fig:fd-method}
\end{figure}

The standard approach to hydrodynamic simulation is to apply the finite-difference (or finite-volume) method to a set of partial differential equations (PDE) of multi-phase flow through a porous medium. Let $\theta$ represent static reservoir variables (computational grid, initial permeability and porosity fields, etc.).
Reservoir state at time $t$ 
will be denoted as $s(t)$ and contain pore pressure, gas content, oil, water and gas saturations.
Let also denote production and injection schedules at time $t$ as $u(t)$ and call them control variables. In the classical simulation, the control variable can be defined in many ways. We will assume it contains bottomhole pressures for all production wells and injection rates for all injection wells (at time $t$). The output of the simulation at time $t$ is the state at the next time step $s(t + \Delta t)$. Thus, one step of the standard simulation process can be described as
\begin{equation}
    s(t + \Delta t) = \textrm{PDE\_solver}(s(t), u(t), \theta, \Delta t) \, .
    \label{simulation}
\end{equation}
The example of classical hydrodynamic simulation with the finite-differences method is shown in Fig.~\ref{fig:fd-method}. Now we move to description of the proposed model.

\subsection{Latent space dynamics}

The proposed reservoir simulation model  is inspired by the Reduced Order Modelling (ROM) technique presented in \cite{drrnn-2018} and recently applied e.g. in \cite{Jin_etal}
and the Neural Ordinary Differential Equations (neural ODEs) introduced in \cite{nde}.
For brevity we call the model
Neural Differential Equations based Reduced Order Model (NDE-b-ROM).
The idea is to translate the reservoir dynamics
into a latent space using encoder-decoder NNs and reconstruct latent space dynamics using differentiable neural ODE. This gives a more flexible approach in contrast to
linear decomposition models (e.g. Dynamic Mode Decomposition \citep{dmd}) and can be compared
to several non-linear models based on NNs proposed in \cite{temirchev, e2c-2015, rce-2017}.

More precisely, we approximate reservoir dynamics in a space of compressed (latent) reservoir state representations $z(t)$. In the following description we will use bold letters (e.g. $\bf g$, $\bf E_
{\theta}$) to indicate functions that are unknown \textit{a priory} and are specified during the model training stage. Of course, after the model is trained, these functions are considered as completely defined. Thus we assume an existence of mappings ${\bf E_s}: \mathcal{S} \rightarrow \mathcal{Z}$ and ${\bf E_z}: \mathcal{Z} \rightarrow \mathcal{S}$ between full-order states and its latent representations such that the composition ${\bf E_z} \circ {\bf E_s}$ is close to the identical operator. Latent space dynamics is assumed to be governed by an ODE of the form:
\begin{equation}
     \frac{dz}{dt} = {\bf g}(z(t), \hat{u}(t), \hat{\theta}) \, ,
\label{latent_dynamics}
\end{equation}
were ${\bf g}(\cdot)$ is some non-linear function, $\hat{u}$ and $\hat{\theta}$ represent latent control and latent static variables of an oilfield. Latent control and static variables are assumed to be obtained from mappings ${\bf E_u}: \mathcal{U} \rightarrow \hat{\mathcal{U}}$ and ${\bf E_{\theta}}: \Theta \rightarrow \hat{\Theta}$ respectively. 

\begin{figure}[h]
\centering
\includegraphics[width=0.85\textwidth]{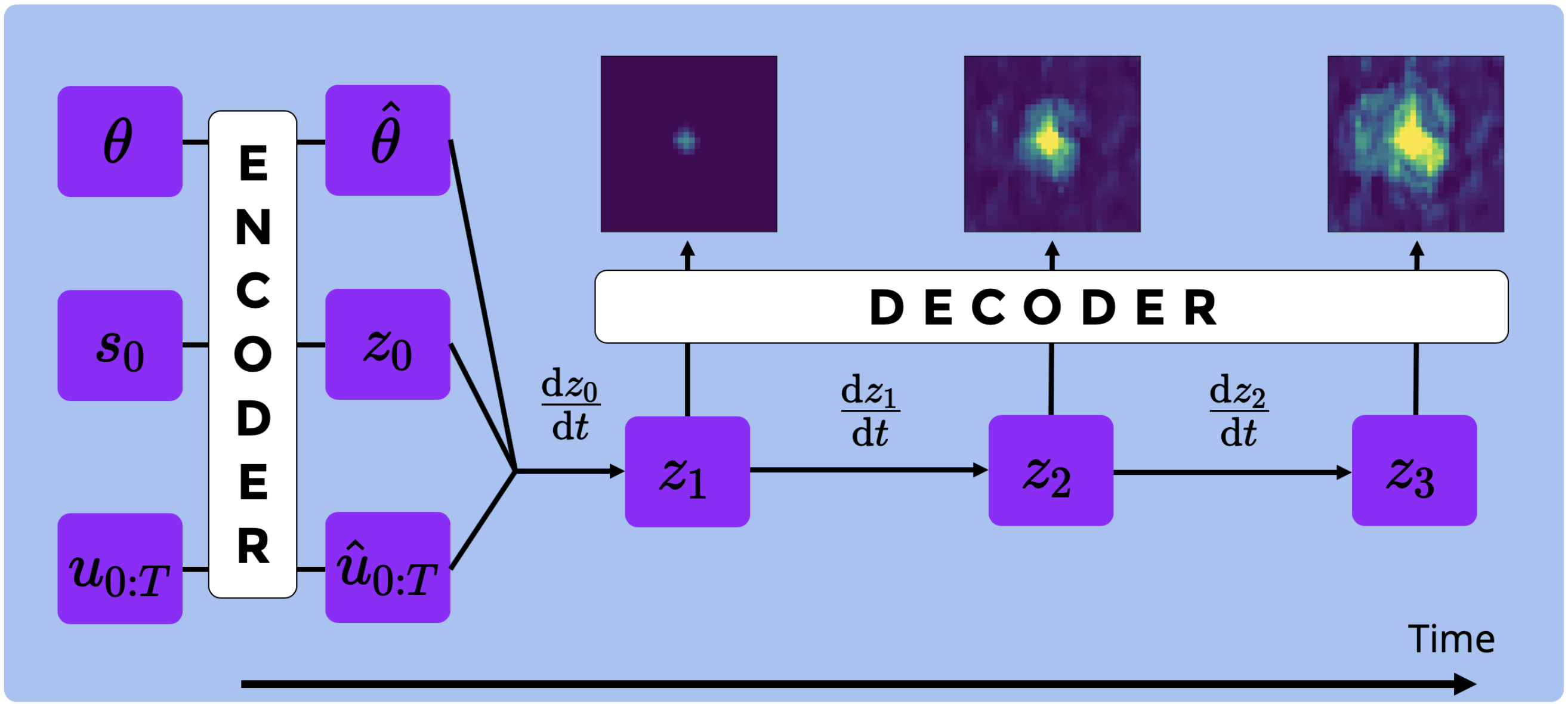}
\caption{Simulation scheme with the NDE-b-ROM model. Here $\theta$, $s_0$ and $u_{0:T}$ denote reservoir static variables, initial state and control parameters for the time interval $(0:T)$. These variables pass through the encoder (which is specific for each variable) and are mapped into the latent space variables denoted as 
 $\hat\theta$, $z_0$ and $\hat u_{0:T}$. Solving the ODE (\ref{latent_dynamics}) in the latent space we obtain a set of latent solutions $z_1$, $z_2$, etc. Decoding latent states $z_i$ we obtain a solution in the initial space and show only some slices of the 3D cubes obtained.}
\label{fig:nde-b-rom}
\end{figure}

The simulation process
starts with an initial reservoir state $s(0)$ and
requires well control schedule $u(t)$ and static information $\theta$. The next reservoir states $s(t)$ are obtained iteratively as follows::
\begin{itemize}
    \item Encode initial state $z(0) = {\bf E_s}(s(0))$, static variables $\hat{\theta} = {\bf E_\theta(\theta)}$ and control $\hat{u}(t) = {\bf E_u}(u(t))$ for all $t$;
    \item Solve the latent ODE for a required period of time using any appropriate numerical scheme. For example, using an explicit integration scheme: $z(t + \Delta t) = z(t) + \Delta t \cdot {\bf g}(z(t), \hat{u}(t), \hat{\theta})$. As a result we obtain the latent solution $z(t)$ for all $t$;
    \item Decode the latent solution: $s(t) = {\bf E_z}(z(t))$ for all $t$.
\end{itemize}
The overall structure of the proposed process is presented in Fig.~\ref{fig:nde-b-rom}.

\subsection{Neural Network Architecture}

The introduced mappings $\bf E_s$, $\bf E_z$ $\bf E_u$, $\bf E_\theta$ and $\bf g$ are 
represented by fully-convolutional NNs \citep{fcn}. 
A benefit of the fully-convolutional architecture is a natural  scalability of the model. In the context of reservoir simulation it allows processing of oil fields of different sizes.

Mappings $\bf E_s$, $\bf E_u$, $\bf E_\theta$ (encoders) are approximated by 4-layer fully-convolutional NNs. The dimensionality reduction  is  controlled by the stride parameter of convolutions. The mapping $\bf E_z$ (decoder) is approximated by a similar NN without strides. Instead, the decoder should increase the dimensionality of a latent variable. It is achieved by the use of a 3-D analog of the Pixel Shuffle method \citep{pixelshuffle} (we call it Voxel Shuffle). The function $\bf g$ is approximated by a simple 2-layer convolutional network. All the modules use Batch Normalization \citep{batchnorm-2015} and Leaky ReLU non-linearity \citep{Maas2013RectifierNI}. This architecture is a result of the compromise between the model depth and ability
to fit into a limited GPU memory when training on
large reservoir models with a large number of timestamps. Of course, there are many internal
parameters in each layer that might require specification. In order to provide
the full reproducibility on the model we open-source
the code of the model as well as any details of 
data processing and model training steps in the GitHub
repository \href{https://github.com/Skoltech-CHR/DeepField}{https://github.com/Skoltech-CHR/DeepField}.

It should be also noted that in contrast to the standard downscaling-upscaling procedures we do not expect substantial information leakage about reservoir heterogeneities in the latent space. The point is that during the model training stage the encoder-decoder pairs are optimized in a way that their composition acts as the identity transform.

\subsection{Training procedure}
\label{training}

NDE-b-ROM is trained end-to-end in a supervised manner. This requires to have a training dataset $\mathcal{D}$ that gives examples of true dynamics evolution.

In supervised learning, dataset comprises a set of pairs $\{ X_i, Y_i \}_{i=0}^{N}$. Here $X_i$ is known information about $i$-th reservoir:
\begin{equation}
X_i = \{ s_i(0), \theta_i, u_i(0), u_i(t_1), u_i(t_2), \dots \} \,,
\end{equation}
while $Y_i$  is a target containing all the information that should be predicted by a model:
\begin{equation}
    Y_i = \{ s_i(t_1), s_i(t_2), s_i(t_3), \dots \}  \,,
\end{equation}
and $N$ is the dataset size.

Commonly, datasets of sufficient size are not available due to technical and commercial subtleties. To overcome this problem we take several hydrodynamic models of real oilfields, randomize them and perform simulation of state (target) variables using classical reservoir simulators. Randomization is made in two steps:

\begin{itemize}
    \item Create new initial state and static variables by adding a small amount of correlated zero-mean Gaussian noise. Create a new bottomhole pressure schedule by sampling from some distribution (see the discussion below). This step generates $X_i$.
    \item Feed the generated initial data into a classical finite difference hydrodynamic simulator and get true dynamics $Y_i$.
\end{itemize}

The choice of good generative schemes is a non-trivial problem and requires a separate investigation. For simplicity, we applied Gaussian randomization for static variables and initial states.
We use correlated noise in order to vary large-scale field
properties rather than to simulate even larger uncertainties about values
in particular grid cells.
For bottomhole pressure a hand-crafted scheme was used:
\begin{equation} u_i(t) = \varepsilon_0 \frac{1 - sin(\varepsilon_1 t + \varepsilon_2)}{2} \exp(- \varepsilon_3 t) + \varepsilon_4 + \varepsilon_5(t) \, .
\end{equation}
Here $\{ \varepsilon_i \}_{i=0}^{5}$  are random numbers from a uniform distribution; $\varepsilon_5$ is the only component which is resampled at each time step.

\begin{figure}[h]
    \centering
    \includegraphics[width=.6\linewidth]{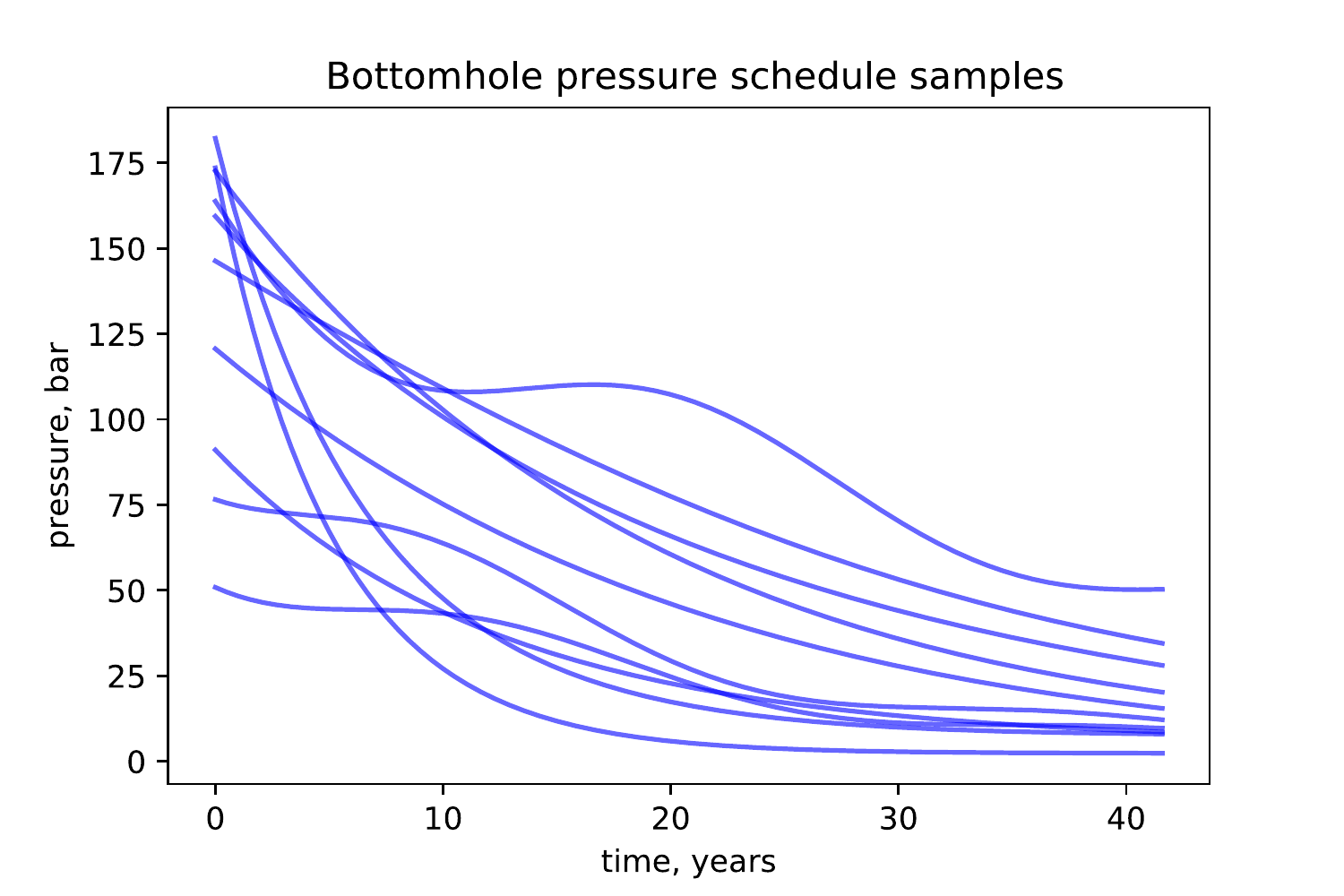}
    \caption{Example of the bottom-hole pressure schedule samples from the generative model.}
    \label{fig:generative}
\end{figure}

Proposed generative scheme gives us realistic initial and boundary conditions as well as a bottomhole pressure schedule (see Fig.~\ref{fig:generative}).

\section{Rates module}
Inflow from the cell connected with perforated well segment is calculated on the basis of standard relation between inflow and pressure drop:
\begin{equation}
    q_P^l(t) = C^l(t)m^l_P(t)\left(p^l(t) - p^l_{con}(t)\right)\,.
\label{rates_model}
\end{equation}
Here $q_P^l(t)$ is a volumetric inflow of a phase $P$ from the $l$-th cell to the well at time $t$, $C^l$ denotes connection productivity index, $m^l_P$ is total mobility of the phase $P$, 
$p^l$ -- pressure in the $l$-th cell, $p^l_{con}$ -- pressure on the well-cell interface.

In case of well parallel to one of the coordinate axis, connection productivity index are calculated explicitly. 
First we find the effective value of the product of formation permeability $K$ and formation thickness $h$:
\begin{equation} \label{eq:kh}
    (Kh)^l = \sqrt{k^l_1 k^l_2}h^l\;,
\end{equation}
here $k^l_1$, $k^l_2$ are permeabilities in the directions perpendicular to the well, $h^l$ is length of perforated well segment in the $l$-th cell. Then for the connection  index
we have
\begin{equation}
    C^l = \frac{2\pi (Kh)^l}{\mathrm{\ln}(r^l_o/r_w)}\;,
\end{equation}
where $r_o$ is equivalent radius (Peaceman radius, \cite{peaceman_interpretation_1978}), $r_w$ is a well radius. Equivalent radius $r_0$ is calculated as
\begin{equation}
  r^l_0 = 0.28\frac{\left((d^l_1)^2 \left(\frac{k^l_2}{k^l_1} \right)^{\frac{1}{2}} + (d^l_2)^2 \left(\frac{k^l_1}{k^l_2} \right)^{\frac{1}{2}}\right)^\frac{1}{2}}{\left(\frac{k^l_2}{k^l_1} \right)^{\frac{1}{4}} + \left(\frac{k^l_1}{k^l_2} \right)^{\frac{1}{4}}}\;,
\end{equation}
where $d^l_1$, $d^l_2$ denote sizes of $l$-th block in directions perpendicular to the well.

If a well has arbitrary trajectory it is approximated as a piecewise linear one; for each linear segment projection onto coordinate axis are calculated and
connection productivity indices are calculated separately for each projections. Thus we obtain projections of the connectivity index onto coordinate index 
$C^l_x$, $C^l_y$, $C^l_z$. To obtain a resulting connectivity index for the cell we summarize these indices:
\begin{equation}
    C^l = C^l_x + C^l_y + C^l_z.
\end{equation}

The described scheme of rates calculation as well as the dynamics modules are implemented using PyTorch framework \citep{NEURIPS2019_9015}. This enables automatic gradient propagation through rate calculations and makes it suitable for optimization problems such as history matching.

\section{Adaptation scheme}

Implementing
reservoir simulation model as an end-to-end differentiable neural network, any set of model parameters can be considered as a space for adaptation. Moreover,
standard adaptation goal (minimization of a difference
between simulated and observed data) can be naturally 
extended with regularization terms that, e.g., penalize
material-balance violation or correction amplitudes.
A detailed investigation of various sets of
parameters in combination with regularization terms
should be a matter of separate research, in this paper
we present rather proof-of-concept results and discuss
further research options.

Following a common HM approach we consider adaptation in the space of
rock parameters (porosity and permeability) and extend it with the auxiliary space of connection productivity indices. Gradient
backpropagation through the neural network model allows
sensitivity estimation for each individual grid cell block. To avoid the undesired overfitting, we require that changes
in a cell block should be correlated with neighboring blocks and penalize large amplitudes using $L_2$ regularization. Up to some extent this regularization hinders the capability of the neural network to model various faults. However, providing the model
with an additional 3D tensor describing the distribution of fault should help to take this information into account. We attribute this investigation to future research.

Technically, we split initial grid into small
cubes of four cell blocks in each direction (of course, one can vary cube sizes
to perform adaptation at different spatial scales). Each cube
attributes to a single additive rock correction factor, initialized with small-amplitude zero mean random noise (we found this initialization works better than constant zero initialization). These correction factors will be adjusted during HM and propagate back to cell blocks of the initial grid through bilinear upsampling.

To include connectivity indices in a space of
adaptation parameters we multiply (\ref{rates_model})
by additional connectivity correction factors. 
In order to ensure that connectivity correction factors remain non-negative during adaptation, we will vary its
logarithms instead of the connectivity correction factors itself. Logarithms
are initialized again with small-amplitude zero mean random noise.

The adaptation process works as follows. We iteratively pass
the adaptation time interval with time steps of a fixed size.
At each step, we calculate predicted production rates
and calculate a loss function that penalties a difference
between predicted and target values.
Based on the loss function value, we compute gradients
with respect to rock and connectivity correction factors and accumulate the gradients.
When the time steps reach the end of the adaptation interval,
correction factors are updated according to the accumulated
gradients and the Adam optimization scheme \citep{Adam}.
Then the gradients are set to zero, and the next
iteration begins.

\begin{figure}[h]
    \centering
    \includegraphics[width=.95\linewidth]{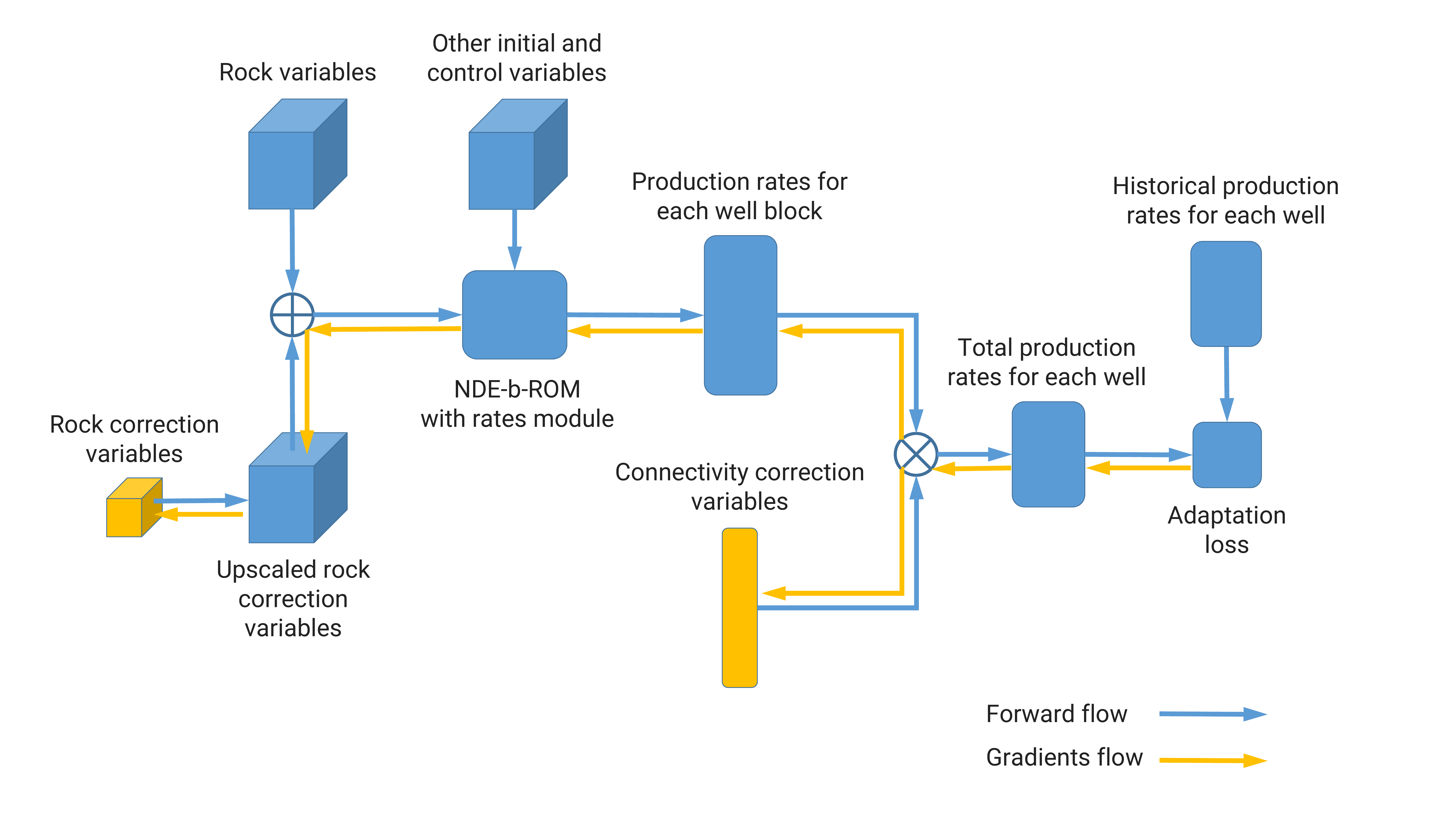}
    \caption{Adaptation scheme. Orange boxes indicate adaptation variables. Rock correction variables are upscaled to the shape of initial rock variables and are added to them. Then the NDE-b-ROM model computes production rates for each
    well block. Production rates are multiplied by 
    connectivity correction factors. Then total
    production rates are compared with historical values and loss function is computed. Using the loss function value gradients are propagated back to adaptation variables. Blue arrows show forward
    data flow, orange arrows show gradients flow.}
    \label{fig:adaptation}
\end{figure}

Total loss function at each iteration
is defined as the aggregated loss over all steps.
Iterations stop when the total loss stops to decrease
substantially. Running the model several times and
varying parameters of the Adam optimization algorithm, we find that increasing of learning rate to 0.3 provides
better and faster convergence. Also, the weight decay
parameter is set to $5\times10^{-4}$, which penalizes large amplitudes of
correction factors.

\section{Reservoir model}
\begin{figure}[h]
    \centering
    \includegraphics[width=0.6\textwidth]{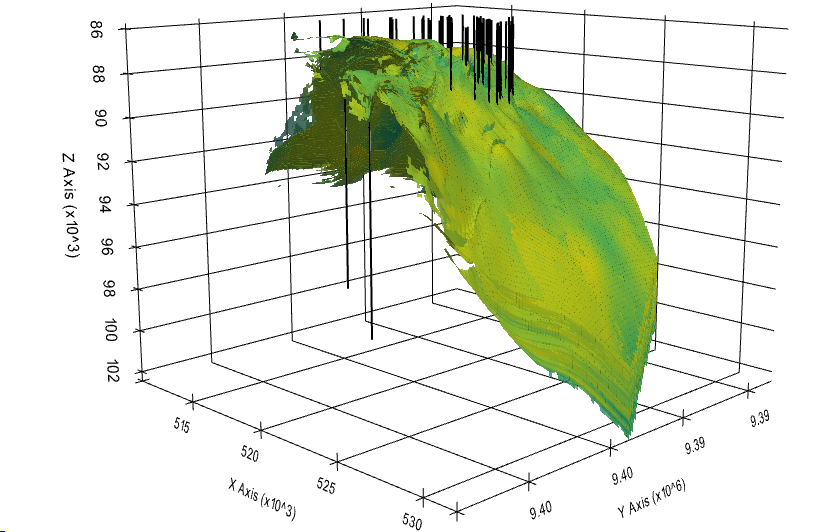}
    \caption{Porosity distribution in the reservoir model.}
    \label{fig:porosity}
\end{figure}

\begin{figure}[h]
    \begin{subfigure}{0.45\textwidth}
        \centering
        \includegraphics[width=\textwidth]{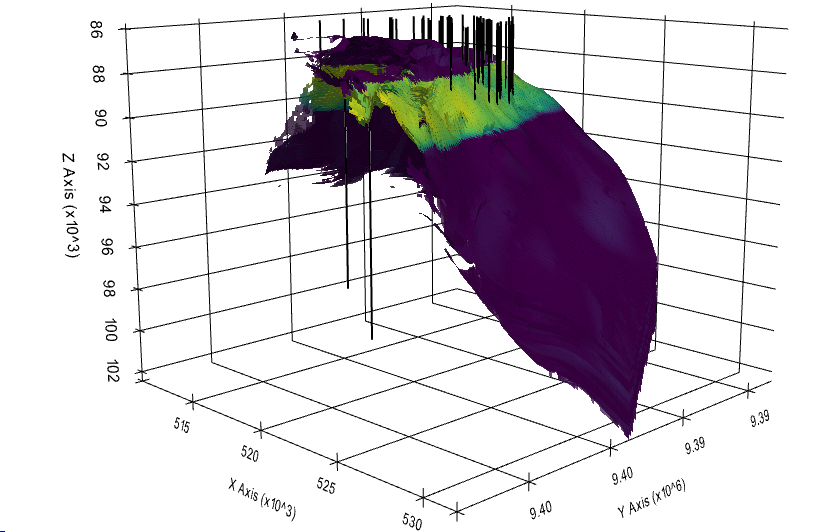}
        \caption{}
        \label{fig:saturation:soil}
    \end{subfigure}
        \begin{subfigure}{0.45\textwidth}
        \centering
        \includegraphics[width=\textwidth]{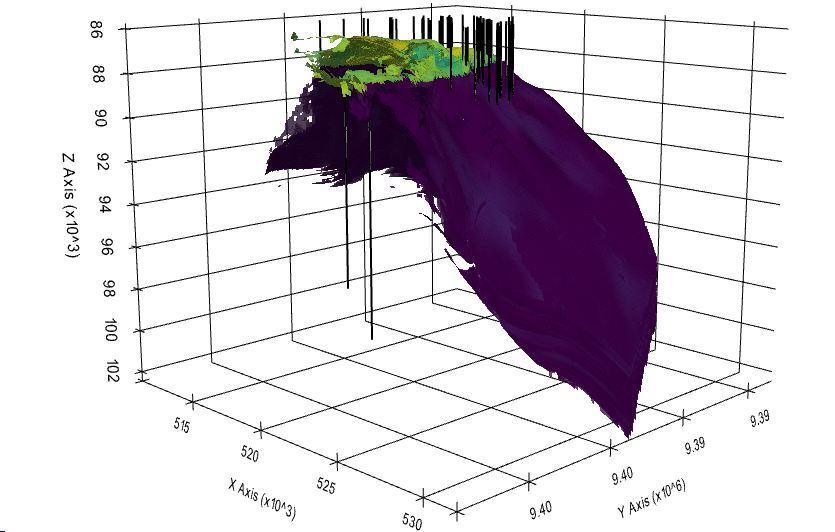}
        \caption{}
        \label{fig:saturation:sgas}
    \end{subfigure}
    \caption{\ref{sub@fig:saturation:soil} -- oil saturation distribution, \ref{sub@fig:saturation:sgas} -- gas saturation distribution in the reservoir model.}
    \label{fig:saturation}
\end{figure}

For numerical experiments we used a synthetic model of the one of Western Siberia oilfield.
The model grid is represented in corner point geometry with $145\times121\times210$ cells, about 1.3M of which are active. In the Fig.~\ref{fig:porosity} structure of the model as well as porosity distribution are presented. Fig.~\ref{fig:saturation} shows oil and gas saturation distributions.

The oilfield has anticlinal shape with gas cap in the top and oil fringe under the gas cap. The complex structure of the model with both gas cap and underlying water make it very sensitive to modelling accuracy. The simulator has to be capable of proper simulation of water and gas coning effects.

\begin{figure}[h]
\centering
\includegraphics[width=0.75\textwidth]{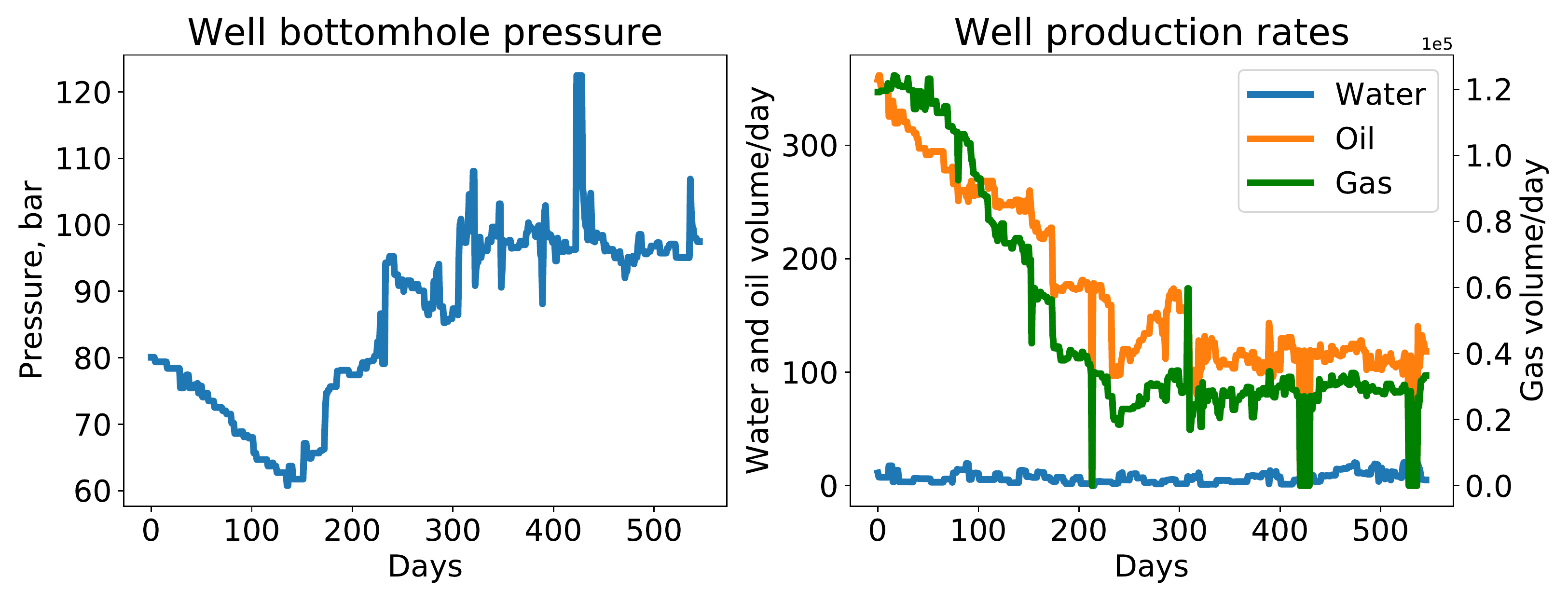}
\caption{Historical recorded bottomhole pressure (left panel) and
production rates (right panel) for
a sample well. Note the double axis (one for water and oil volumes and one for gas volume) in
the right panel.}
\label{fig:production_history}
\end{figure}

The hydrocarbons are recovered with the use of 64 production wells located in both oil and gas areas. Due to the early stage of reservoir recovery, only a small fraction of wells has more or less complete production history. 
We carefully selected a time interval and
a set of wells involved in this research to eliminate low-quality
records and, in particular, to eliminate wells with only
fragmentary history available. The reason for this preprocessing is as follows. It is clear that data quality is
essential for accuracy of adaptation results. However, it is rather
difficult to separate the impact of data quality and model capability itself. The aim of this paper is to demonstrate the generic
approach, while application to different reservoir models with
specifically complicated history data might require a
customization of data preprocessing steps.
The latter discussion is out of the scope of this paper.
Finally, we use a set of 12 wells
and a time interval of 1.5 years.
Each well has daily recorded historical oil, water and gas  production rates as well as bottomhole pressure.
The recorded bottomhole pressure is used as control parameter in reservoir simulation.
An example of recorded history for one of the wells is presented in the Fig.~\ref{fig:production_history}.

\section{Results}

In this section, we provide results of an experiment in which rock
parameters are varied on a grid downsampled by factor 4
with respect to original grid sizes. Note that downsampling
is applied not due to resource limitation but as a natural
regularization for HM. The loss function for HM is defined
as mean squared error (MSE) between predicted and historical (target) rates
for each well  and aggregated over all wells and fluid phases
(gas, water, oil). To normalize substantially different
scales in production rates of various fluid phases, we
apply logarithmic calibration before computing the MSE.
Also, we apply a linear time-weighting function that
increases an impact of error with time progressing. 
An intuition behind this weighting is that errors 
in recent rates are more important in comparison
to more time-remote errors.

\begin{figure}[h]
\centering
\includegraphics[width=0.5\textwidth]{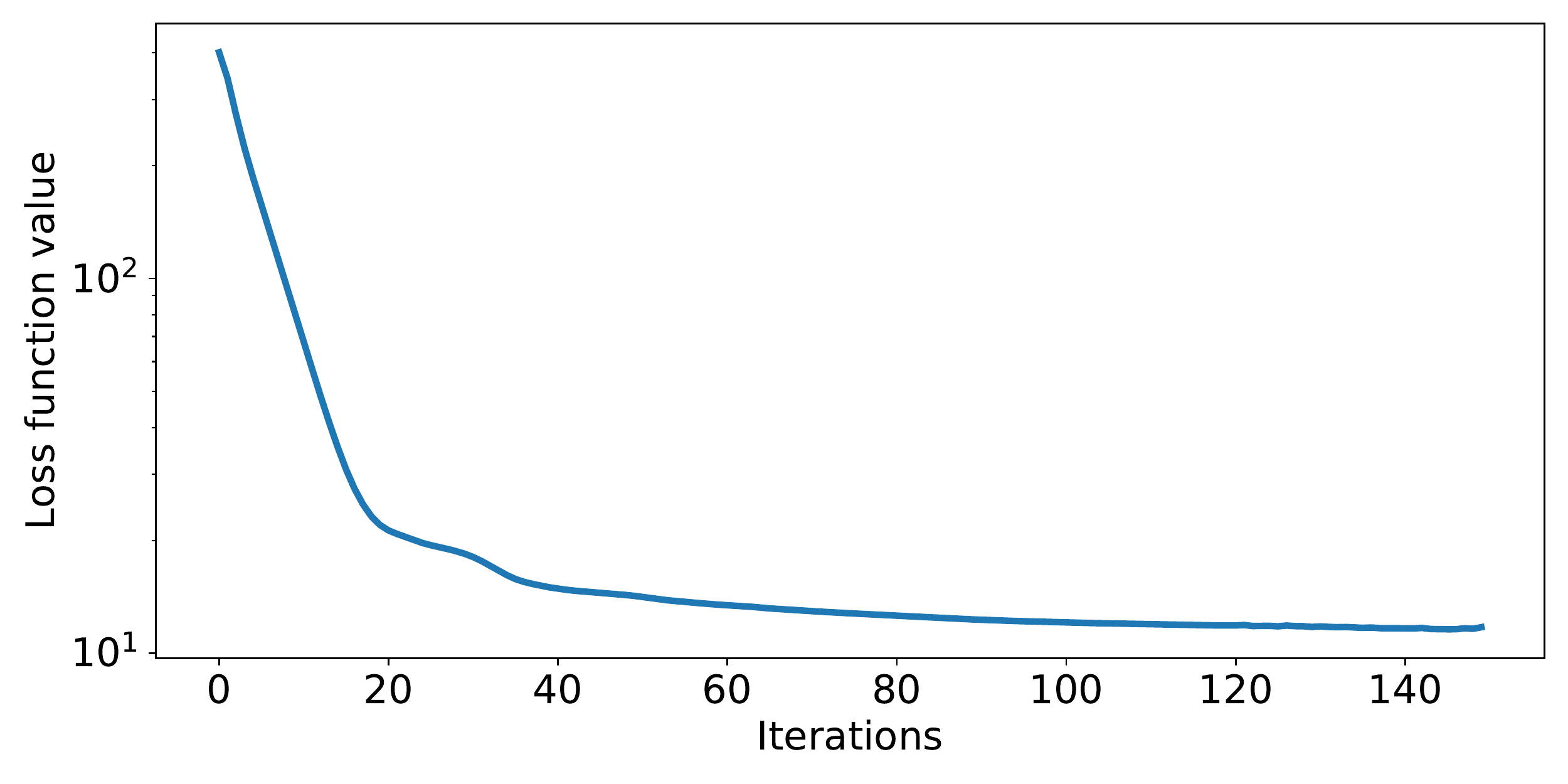}
\caption{Adaptation loss function decay against iterations.}
\label{fig:loss}
\end{figure}

Fig.~\ref{fig:loss} shows the total
loss function decrease against iterations. After 150
iterations loss stops to decrease and begins to fluctuate near a constant value. We stop the adaptation process at this moment.

\begin{figure}[h]
\centering
\includegraphics[width=0.8\textwidth]{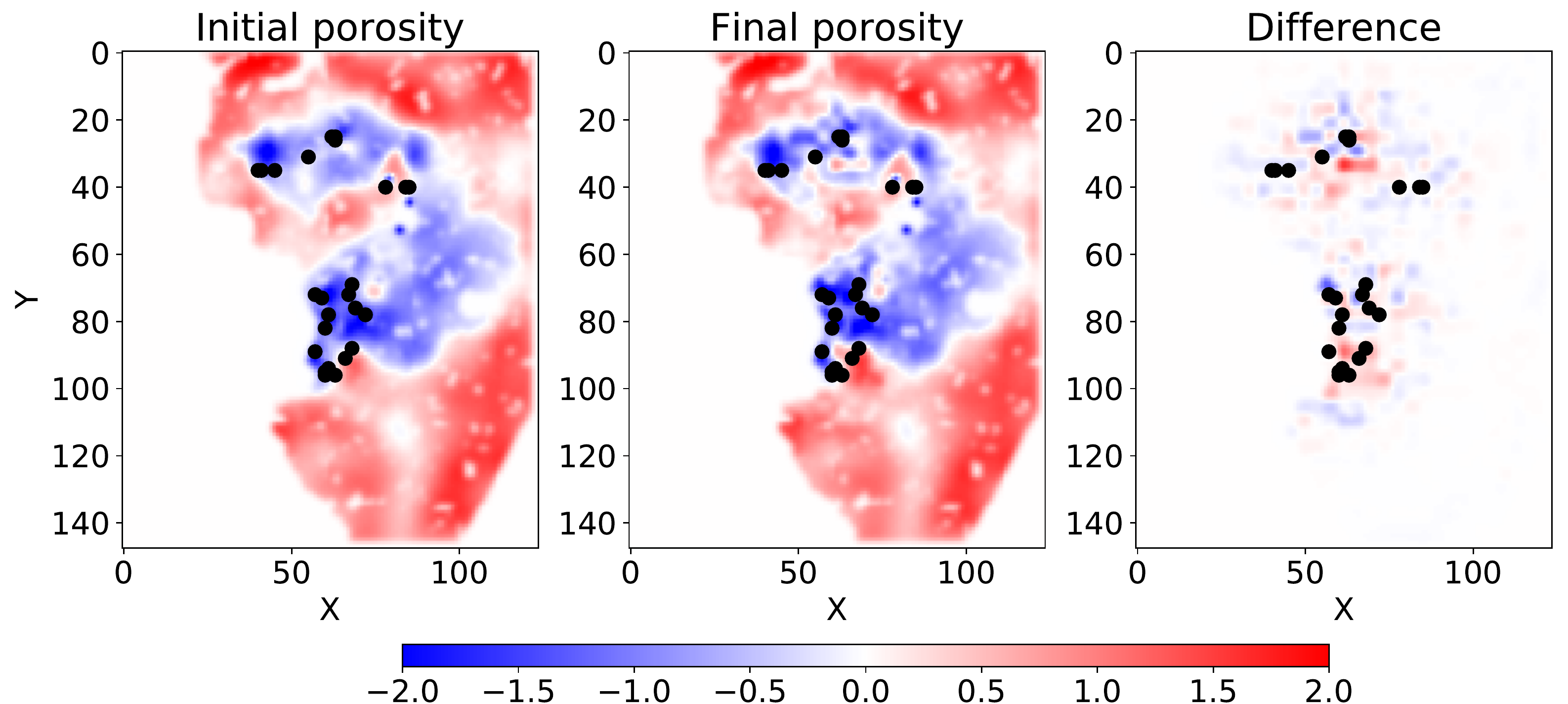}
\caption{Sample horizontal slice of the cube of normalized porosity values. The left panel shows the initial porosity distribution. The middle panel is the result of adaptation. The right panel
shows the difference between the plots. Black dots indicate
the location of production wells.
Weights regularization is applied during adaptation.}
\label{fig:poro_slice}
\end{figure}

\begin{figure}[h]
\centering
\includegraphics[width=0.8\textwidth]{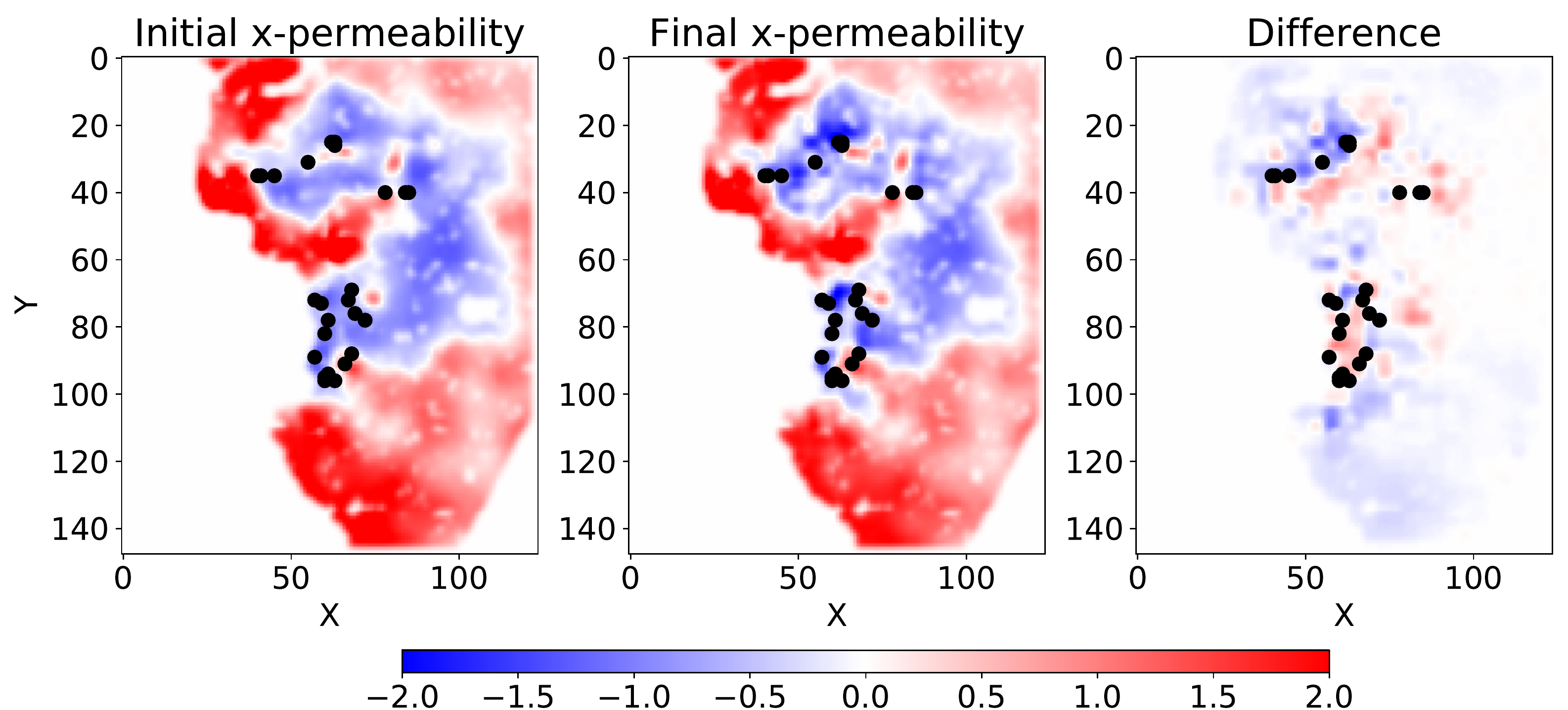}
\caption{Same as Fig.~\ref{fig:poro_slice} but for the cube of normalized x-permeability.}
\label{fig:perm_slice}
\end{figure}

\begin{figure}[h]
\centering
\includegraphics[width=0.8\textwidth]{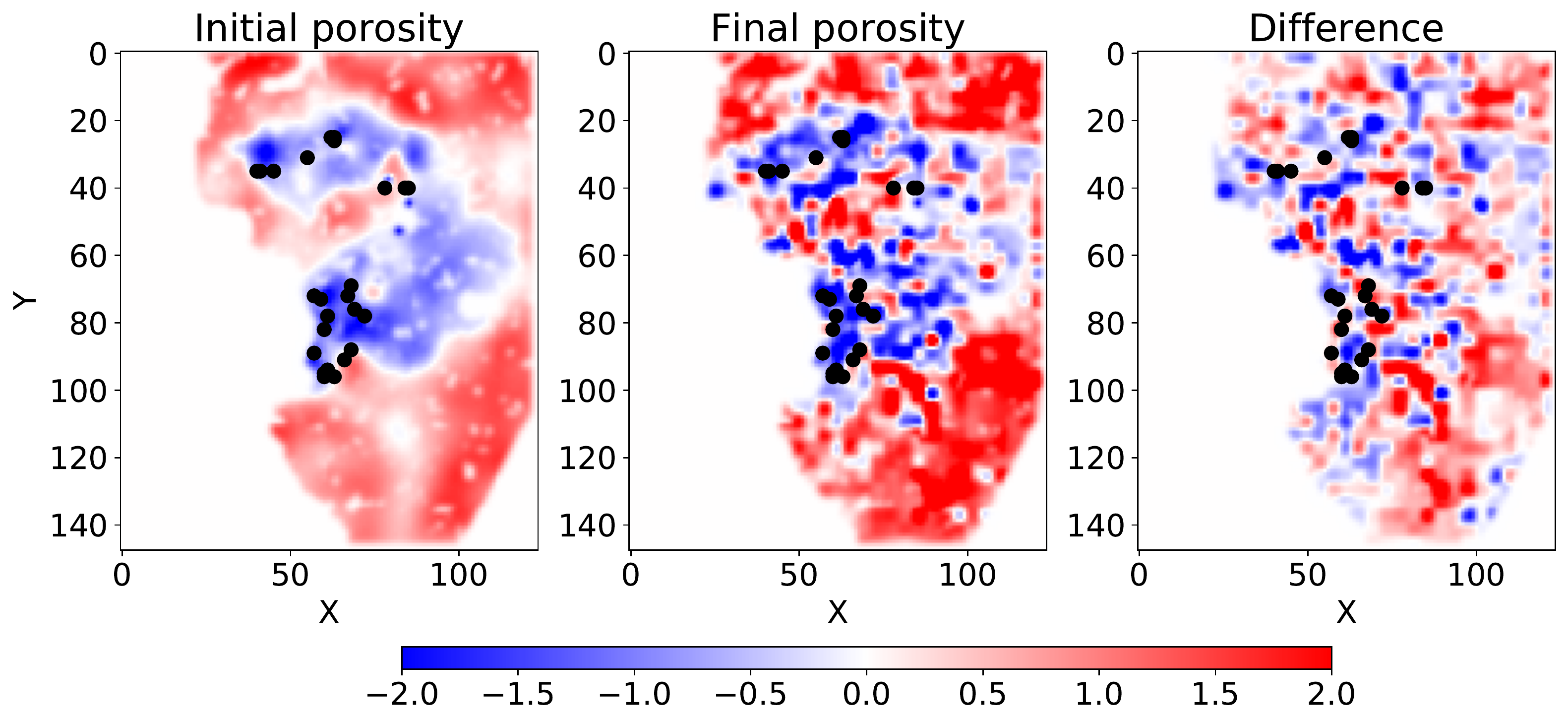}
\caption{Same as Fig.~\ref{fig:poro_slice} but without weights regularization during adaptation.}
\label{fig:poro_slice_no_reg}
\end{figure}

Fig.~\ref{fig:poro_slice} and  Fig.~\ref{fig:perm_slice} show a sample horizontal slice 
of the cube of normalized porosity and
x-permeability. Note that due to weight regularization the adaptation process affects only
a small region around the production wells. In contrast,
a model without regularization makes changes even in areas remote from production wells, which is less physically
sound (see Fig.~\ref{fig:poro_slice_no_reg} for comparison).

The next Fig.~\ref{fig:corr} shows how much
are the changes in adaptation parameters
introduced by the HM. On average, rock variables obtain a small negative bias -0.01 (significant statistically). However, we find that the
initial phase content (i.e.
porosity multiplied by phase saturation and cell volume) changes only by
less than $1\%$.
Quite interesting, we find that almost all connectivity
correction factors are distributed near 0 and 1 (being unit initialized). Since 
the connectivity correction factor is multiplicative, value 1 means no correction is applied, while 0 corresponds to a effectively closed cell's perforation. Note that this result requires a separate detailed investigation in order
to avoid physically irrelevant situations when e.g. several well's block are substituted with a single block of increased connectivity index. We admit that additional 
regularization terms might be proposed to control the
distribution of connectivity correction factors.

\begin{figure}[h]
\centering
\includegraphics[width=0.6\textwidth]{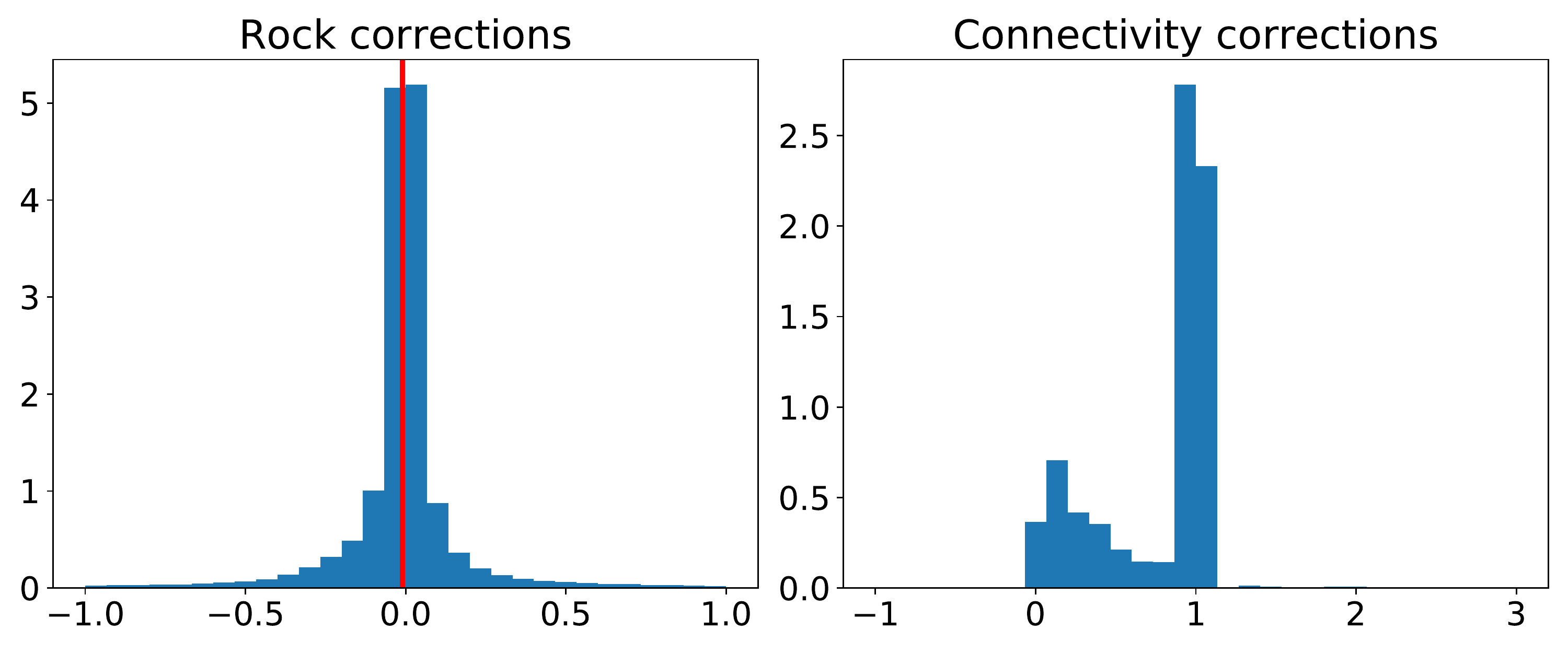}
\caption{Distribution of rock (left panel) and 
connectivity (right panel) correction factors. 
Note that rock corrections are additive,
while connectivity corrections are multiplicative. Red line in the left plot
shows the mean value of the distribution.}
\label{fig:corr}
\end{figure}

\begin{figure}[h]
\centering
\includegraphics[width=0.8\textwidth]{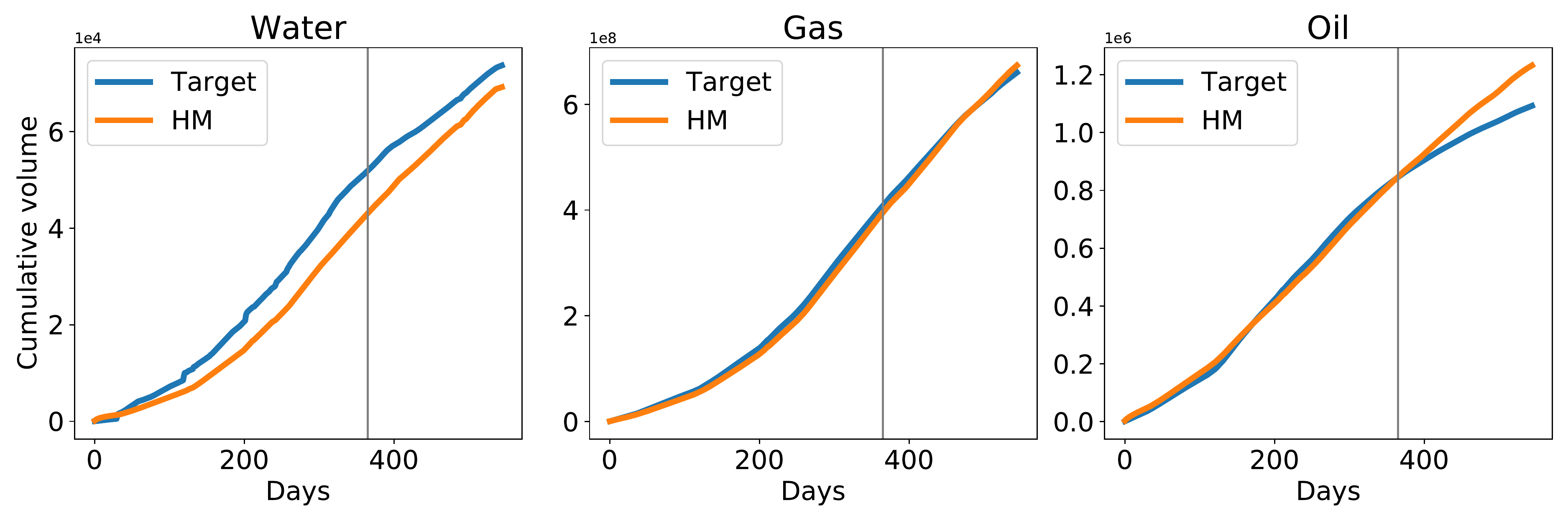}
\caption{Cumulative production rates over all wells. Blue line shows historical values, orange line is a model simulation. Vertical line separates 
adaptation and prediction periods.}
\label{fig:HM_all_wells_rates}
\end{figure}

\begin{figure}[h]
\centering
\includegraphics[width=0.8\textwidth]{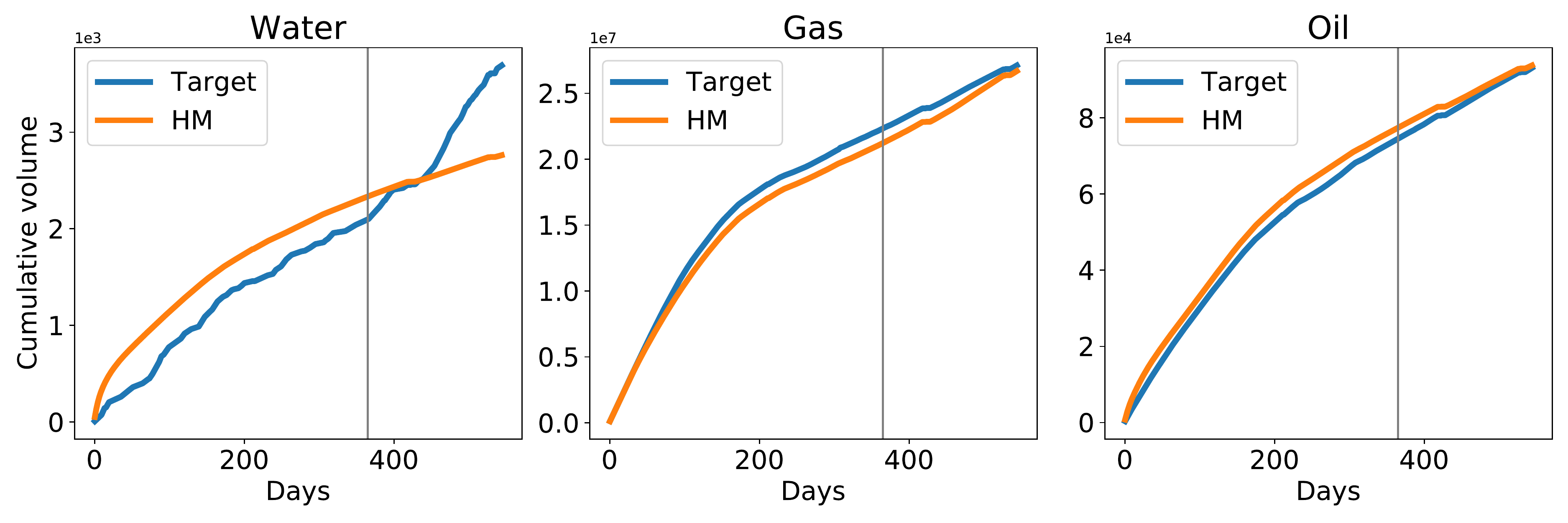}
\caption{Cumulative production rates for a sample well. Blue line shows historical values, orange line is a model simulation. Vertical line separates 
adaptation and prediction periods.
Note that daily
historical rates and borehole pressure for the same
well are shown in Fig.~\ref{fig:production_history}.}
\label{fig:HM_one_wells_rates}
\end{figure}

Fig.~\ref{fig:HM_all_wells_rates} shows
a comparison of target and simulated cumulative production rates. Note that the time interval is
split into two parts. The first one
shows a comparison within the adaptation period. The second one demonstrates a prediction against historical values. We observe that the model is possible to reproduce the historical values
given in the adaptation period and can be
used for forecasting on an interval that is at least half of the adaptation period length.
The same plot but for a sample well is shown in Fig.~\ref{fig:HM_one_wells_rates}.
We observe that predicted values partially go off-track, e.g. for the water. We address this issue to the current limitation of the neural network model that supports only the limited scope of features and events given in the reservoir history data. More detailed technical description is available in the documentation that supports the open-sourced code.

Fig.~\ref{fig:corr} shows a correlation diagram
between predicted and target cumulative production rates over all wells. We observe that
for each phase (water, gas, and oil), the correlation coefficient ($R$ value) is 0.94 or above. This indicates that the adaptation process successfully matches the production rates
of individual wells.

\begin{figure}[h]
\centering
\includegraphics[width=0.8\textwidth]{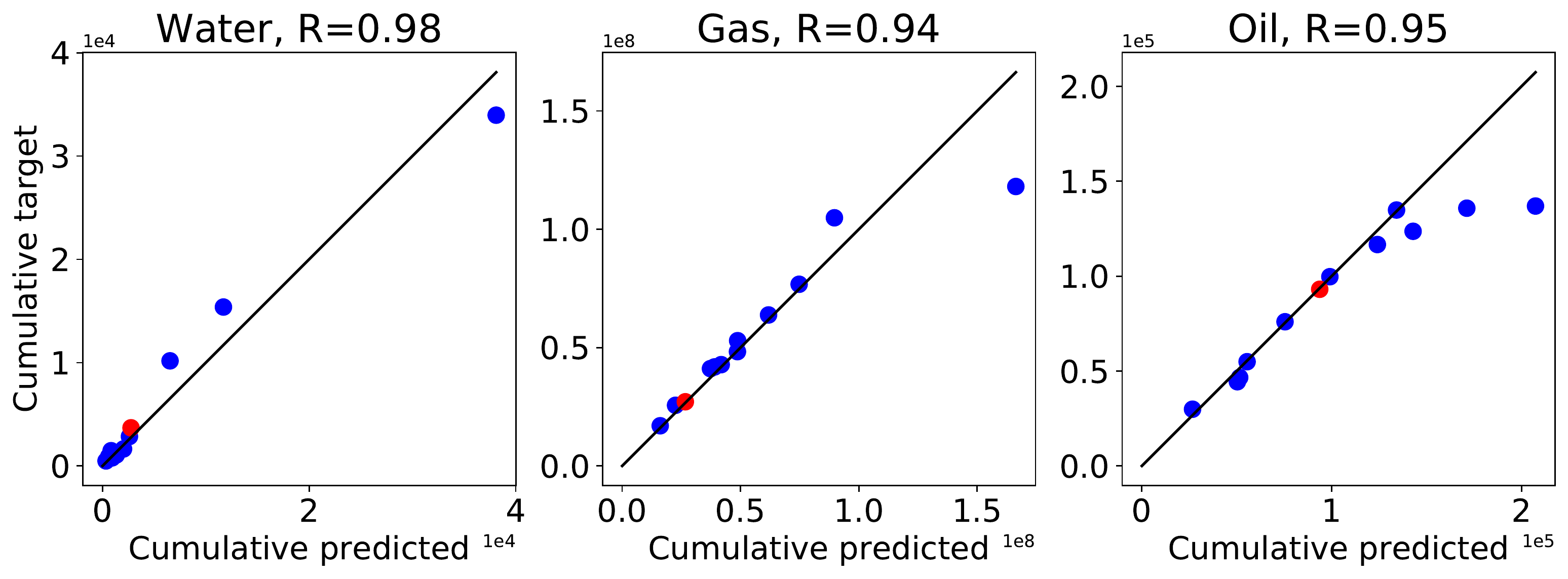}
\caption{Correlation between target
and predicted cumulative production rates for individual wells (shown as dots).
Correlation coefficients are given
in the title of each plot.
Inclined gray line corresponds to $R=1$.
Well shown in red is the same well as in Fig.~\ref{fig:HM_one_wells_rates} and
Fig.~\ref{fig:production_history}.}
\label{fig:corr}
\end{figure}

\begin{figure}[h]
\centering
\includegraphics[width=0.4\textwidth]{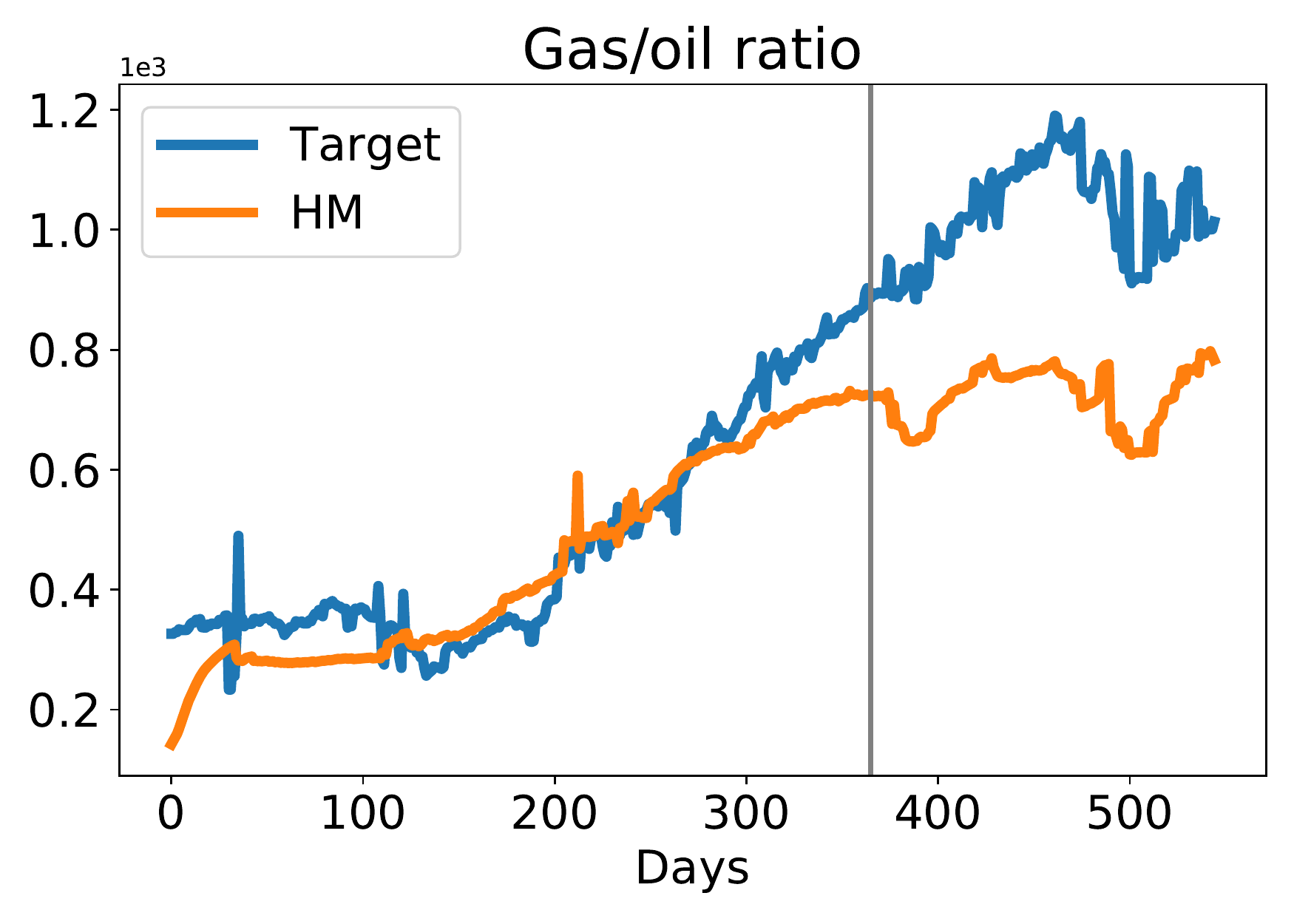}
\caption{Gas/oil ratio according to daily gas and oil
production rates. Blue line shows target (historical)
values, orange line is a simulation. Vertical line
separates adaptation and prediction intervals.}
\label{fig:HM_gas_oil_ratio}
\end{figure}

In Fig.~\ref{fig:HM_gas_oil_ratio} we provide a gas/oil
a ratio computed according to daily simulated and historical gas and oil production rates. Since the 
gas/oil ratio is a critical parameter in the reservoir
recovery management, we find that predicted values
are in partial agreement with actual historical values.

\begin{figure}[h]
\centering
\includegraphics[width=0.8\textwidth]{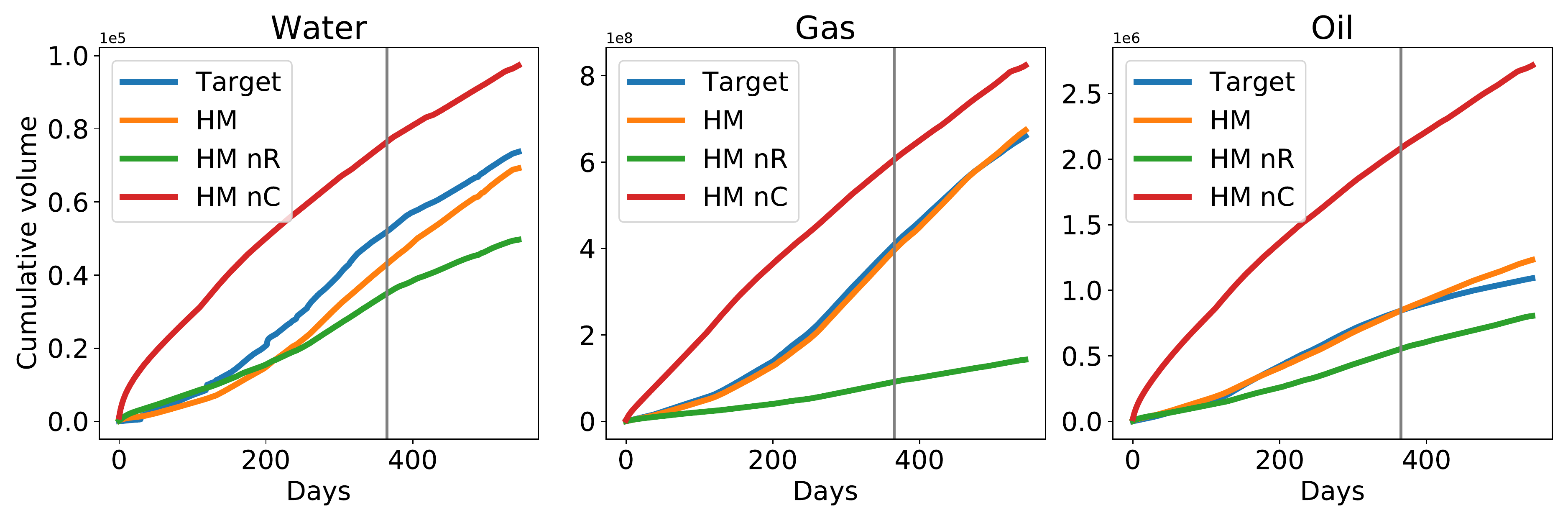}
\caption{Cumulative production rates aggregated over all wells. Target (blue line) shows
historically recorded values. The orange line (labeled "HM")
is a model simulation. The green line (labeled "HM nR") shows
a model simulation where rock correction factors are not applied. The red line (labeled "HM nC") shows
a model simulation where connectivity correction factors are not applied.}
\label{fig:HM_nRC}
\end{figure}

To demonstrate the role of rock and connectivity correction factors, we exclude from simulation either rock or connectivity correction factors
and compare the simulation with target values and a simulation where both factors are included.
One can note in Fig.~\ref{fig:HM_nRC}
that each set of adaptation variables is
meaningful, and its combination gives substantially better results.

\begin{figure}[h]
\centering
\includegraphics[width=0.8\textwidth]{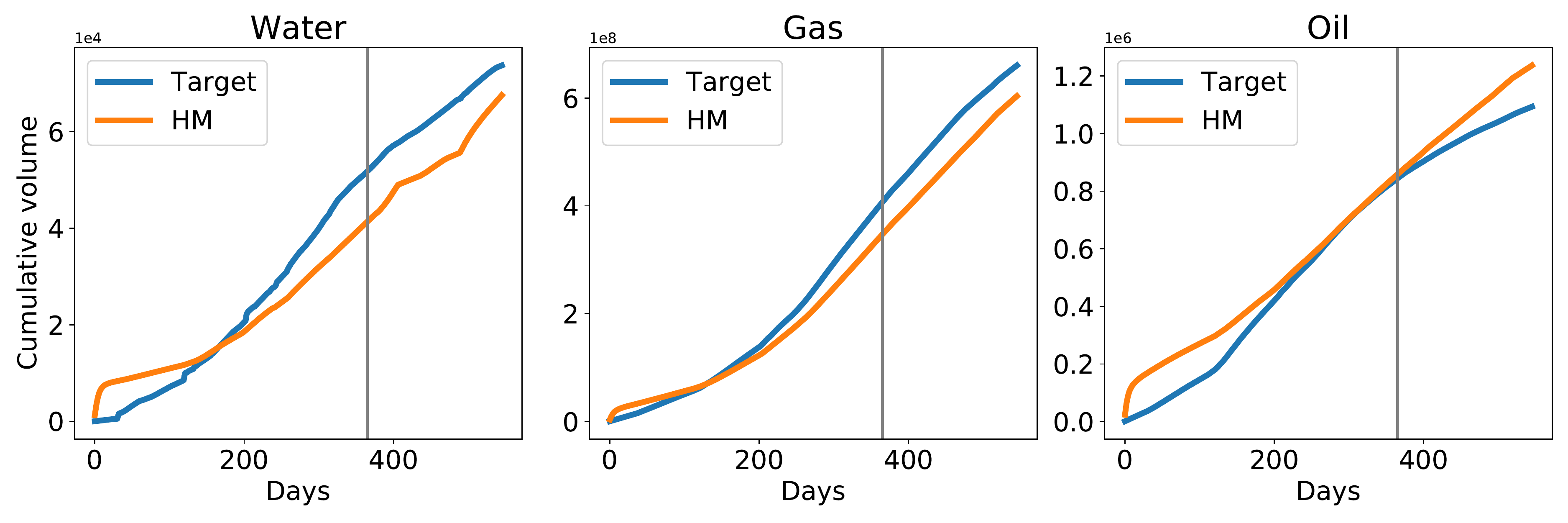}
\caption{Same as Fig~\ref{fig:HM_all_wells_rates} but only rock correction factors were
varied during adaptation.}
\label{fig:HM_all_wells_rates_rock}
\end{figure}

\begin{figure}[h]
\centering
\includegraphics[width=0.8\textwidth]{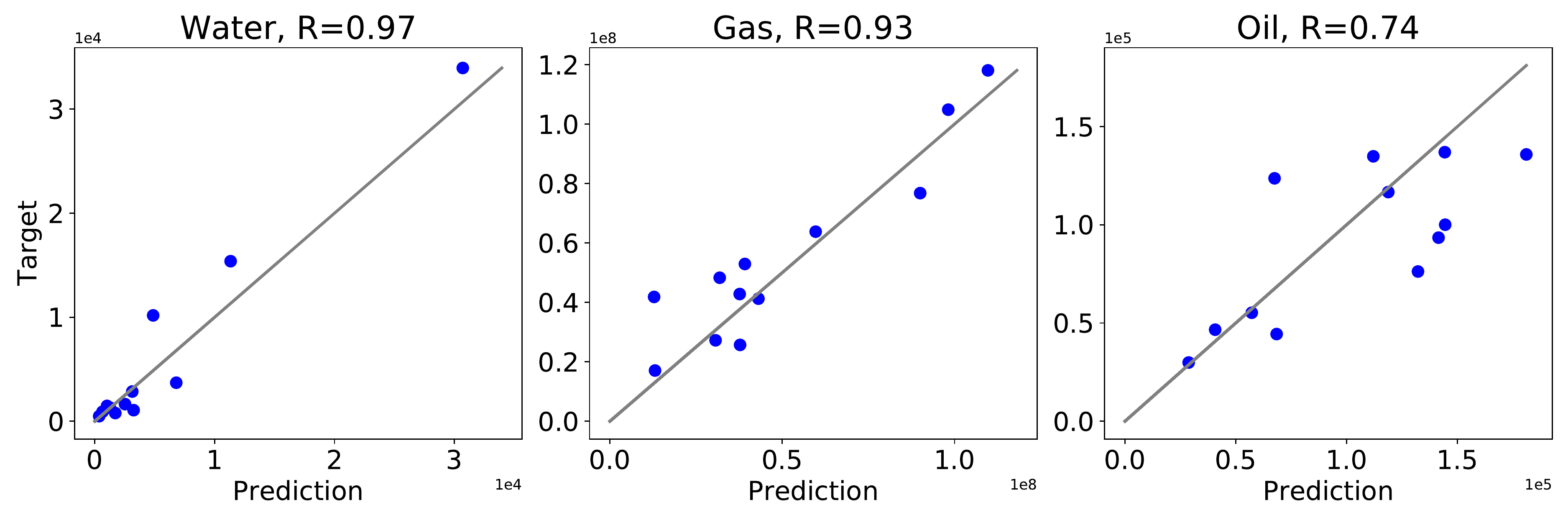}
\caption{Same as Fig~\ref{fig:corr} but only rock correction factors were
varied during adaptation.}
\label{fig:corr_rock}
\end{figure}

Finally, we demonstrate that adaptation only in the space of rock correction factors (without including connectivity factors) limits
the model quality. Indeed, while 
cumulative production rates over all wells shown in  
Fig.~\ref{fig:HM_all_wells_rates_rock}
are compatible  with previous Fig.~\ref{fig:HM_all_wells_rates},
correlation between individual wells become substantially worse (compare  Fig.~\ref{fig:corr_rock} and Fig.~\ref{fig:corr}). We conclude 
that adaptation in the joint space of 
rock and connectivity factors
allows better 
matching for individual wells. 
One can also conclude from the last example that for evaluation of
different adaptation models total
production rates can not be a single
benchmark and additional metrics should be considered as well.

\section{Conclusions}

We presented an end-to-end neural network approach that allows reservoir simulation and history matching with standard gradient-based optimization algorithms. The neural network model has  initial geological parameters of the 3D reservoir model in the input. In the output, the model returns wells' production rates.
By construction, neural network models allow gradients propagation to any internal and input variables. 
Using a dataset of development scenarios, we train the neural network to simulate general geological relations. In this case, internal network variables are optimized. Using historical records on real production rates and bottomhole pressure, we solve the history matching problem. In this case, 
initial rock parameters and connectivity indices were optimized.
As we demonstrated, the final model
allows reliable simulation of historical production rates and forecasting of reservoir dynamics.

It should be noted that the suggested neural network
approach is not to replace standard industrial reservoir
simulation software. The goal is to obtain a 
substantially faster simulation tool, probably at the cost of acceptable accuracy decrease. In this research, we consider the reservoir model of about 3.7M total grid cell size and about 1.3M of active cells. Simulation of daily production rates for 1.5 years time interval takes about 1 minute (using modern GPU workstation). Note that the computation also includes a complete simulation of pressure and phase saturation cubes. This result is several orders of magnitude faster
in comparison to current industrial reservoir
simulation software. The adaptation process for one-year period takes about 3 hours.

The neural network approach opens a broad and convenient way for implementation 
of many reservoir simulation and adaptation strategies.
The point is that one can easily combine variables
to be optimized during HM. For example, in this research, we consider a joint adaptation in the space of rock parameters and connectivity indices. Also, the HM and
forward simulation problems can be naturally extended
by additional regularization terms, including
a control for proper conservation of physical parameters such as mass of components. 
Investigation and comparison
of the various experiment settings is a matter of future 
research, which looks optimistic taking into account
proof-of-the-concept results demonstrated in this paper. 

This research is completely based on the source-code available in the GitHub repository \href{https://github.com/Skoltech-CHR/DeepField}{https://github.com/Skoltech-CHR/DeepField}.

\section*{Acknowledgement}
We thank the reviewers for valuable comments and suggestions.
The work is supported by Gazprom Neft and the  Ministry of Science and Higher Education of the Russian Federation under agreement No. 075-10-2020-119 within the framework of the development program for a world-class Research Center.

\bibliographystyle{cas-model2-names}

\bibliography{mybibfile}

\end{document}